\documentclass{article}
\usepackage{spconf,amsmath,graphicx,hyperref,subfig}

\title{ROBUST AND FINE-GRAINED PROSODY CONTROL OF \\END-TO-END SPEECH SYNTHESIS}
  \name{Younggun Lee, Taesu Kim}
  \address{Neosapience, Inc., Seoul, Republic of Korea}
\begin{document}
\ninept
\maketitle
\begin{abstract}
We propose prosody embeddings for emotional and expressive speech synthesis networks. The proposed methods introduce temporal structures in the embedding networks, thus enabling fine-grained control of the speaking style of the synthesized speech. The temporal structures can be designed either on the speech side or the text side, leading to different control resolutions in time. The prosody embedding networks are plugged into end-to-end speech synthesis networks and trained without any other supervision except for the target speech for synthesizing. It is demonstrated that the prosody embedding networks learned to extract prosodic features. By adjusting the learned prosody features, we could change the pitch and amplitude of the synthesized speech both at the frame level and the phoneme level. We also introduce the temporal normalization of prosody embeddings, which shows better robustness against speaker perturbations during prosody transfer tasks.

\end{abstract}
\begin{keywords}
Prosody, Speech style, Speech synthesis, Text-to-speech
\end{keywords}
\section{Introduction}
\label{sec:intro}

Since Tacotron \cite{Tacotron} paved the way for end-to-end Text-To-Speech (TTS) using neural networks, researchers have attempted to generate more naturally sounding speech by conditioning a TTS model via speaker and prosody embedding \cite{DV2, Mimic1, Prosody, SToken, text2style}. (We use the term \textit{prosody} as defined in earlier work \cite{Prosody} henceforth.) Because there is no available label for prosody, learning to control prosody in TTS is a difficult problem to tackle. Recent approaches learn to extract prosody embedding from reference speech in an unsupervised manner and use prosody embedding to control the speech style \cite{Prosody, SToken}. These models have demonstrated ability to generate speech with expressive styles with Tacotron \cite{Tacotron} using prosody embedding. They can also transfer the prosody of one speaker to another using a different speaker ID while leaving the prosody embedding unchanged. However, we observed two limitations with the above models.

First, controlling the prosody at a specific moment of generated speech is not clear. Earlier works focused on prosody embedding with a fixed length (a length of 1 in their experiments) regardless of the length of the reference speech or that of the text input. A loss of temporal information when squeezing reference speech into a fixed length embedding is highly likely. Therefore, fine-grained control of prosody at a specific moment of speech is difficult for embedding with a fixed length. For example, we can set the global style as "lively" or "sad," but we cannot control the prosody of a specific moment with fixed-length embedding. Because humans are sensitive to subtle changes of nuance, it is important to ensure fine-grained control of prosody to represent one's intentions precisely.

Secondly, inter-speaker prosody transfer is not robust if the difference between the pitch range of the source speaker and the target speaker is significant. For example, when the source speaker (female) has higher pitch than the target speaker (male), the prosody-transferred speech tends to show a higher pitch than the usual pitch of the target speaker.

In this work, we focus on solving these two problems. We will introduce two types of variable-length prosody embedding which have the same length as the reference speech or input text to enable sequential control of prosody. In addition, we will show that normalizing prosody embedding helps to maintain the robustness of prosody transfers against speaker perturbations. With our methods, speaker-normalized variable-length prosody embedding was able to not only to control prosody at each specific frame, but also to transfer prosody between two speakers, even in a singing voice.


\section{RELATED WORK}
\label{sec:related}

Prosody modeling had been done in a supervised manner by using annotated labels, such as those in ToBI \cite{Tobi}. Problems were reported about hand annotations, and the cost was high \cite{Tobix}.

Skerry-Ryan et al. used convolutional neural networks and a Gated Recurrent Unit (GRU) \cite{GRU} to compress the prosody of the reference speech \cite{Prosody}. The output, denoted by $p$, is fixed-length prosody embedding. They enabled prosody transfers using the prosody embedding, but they could not gain control of prosody at a specific point of time. Another problem was also reported \cite{SToken}; fixed-length prosody embedding worked poorly if the length of the reference speech was shorter than the speech to generate. In addition, variable-length prosody embedding was also implemented using the output of the GRU at every time step \cite{Prosody}. However, this method did not draw attention because it could not obtain satisfactory results given that it was not robust with regard to text and speaker perturbations. We noted the usefulness of variable-length prosody and elaborated on this concept for fine-grained prosody control.

Wang et al. came up with the global style token (GST) Tacotron to encode different speaking styles \cite{SToken}. Although they used the same reference encoder architecture used in earlier work \cite{Prosody}, they did not use $p$ itself for prosody embedding. Using a content-based attention, they computed the attention weights for style tokens from $p$. The attention weights represent the contribution of each style token, and the weighted sum of the style tokens is now used for style embedding. During the training step, each randomly initialized style token learns the speaking style in an unsupervised manner. In the inference mode, it was possible to control prosody by either predicting the style embedding from the reference speech or specifying the attention weights of the style tokens. 
This enables explicit control of the speaking style, but it nonetheless worked only in a global sense. If we are interested in controlling the prosody of a phoneme, it would be ideal to obtain the same prosody for different phonemes when the phonemes are conditioned on the same prosody embedding. However, GST Tacotron generates various types of prosody for input phonemes that are conditioned on the same style embedding, which is not desirable for prosody control. Wang et al. also proposed text-side style control using multiple style embeddings for different segments of input text. This method could roughly change the style of the text segments, but it is limited when used to control phoneme-wise prosody for the reasons mentioned above.

\section{Baseline model}
\label{sec:architecture}

We used a simplified version \cite{Tacotron2} of Tacotron for the base encoder-decoder architecture, but we used the original Tacotron \cite{Tacotron} style of the Post-processing net and the Griffin-Lim algorithm \cite{GLRecon} for spectrogram-to-waveform conversion. For the encoder input $x$, we used the phoneme sequence of normalized text to ease the learning. The one-hot speaker identity is converted into speaker embedding vector $s$ by the embedding lookup layer. Equation \ref{eq:base} describes the base encoder-decoder, where $e, p, \text{and } d$ denote the text encoder state, variable-length prosody embedding, and decoder state, respectively.

\begin{equation}\label{eq:base}
\begin{aligned}
    e_{1:l_e} &= \text{Encoder}(x_{1:l_e})\\
    \alpha_i &= \text{Attention}(e_{1:l_e}, d_{i-1})\\
    e'_i &= \Sigma_j \alpha_{ij} e_j\\
    d_i &= \text{Decoder}(e'_i, s)
\end{aligned}
\end{equation}

Reference speech is encoded to prosody embedding using the reference encoder \cite{Prosody}. A mel-spectrogram of the reference speech proceeds through 2D-convolutional layers. The output of the last convolutional layer is fed to a uni-directional GRU. The last output of GRU $r_N$ is the fixed-length prosody embedding $p$. If we use every output of GRU $r_{1:N}$ for prosody embedding, it forms the variable-length prosody embedding $p_{1:N}$.


\section{Proposed method}
\label{sec:method}

Fine-grained prosody control can be done by adjusting the values of variable-length prosody embedding. We propose two types of prosody control methods: speech-side control and text-side control. Variable-length prosody embedding is used as a conditional input at the encoder module or at the decoder module for speech-side control or text-side control, respectively. In order to do this, we need to align and downsample the prosody embedding to match the length of the prosody embedding $l_p$ with the speech side (the number of decoder time-steps, $l_d$) or the text side (the number of encoder time-steps, $l_e$).

\subsection{Modifications in the reference encoder}
\label{ssec:mod_ref_enc}

We empirically found that the following modifications improved the generation quality. We used CoordConv \cite{coordconv} for the first convolutional layer. According to its construction, Coordconv can utilize positional information while losing the translation invariance. We speculate that the positional information was helpful to encode prosody sequentially. We used ReLU as the activation function to force the values of the prosody embedding to lie in [0, $\infty$].

The proposed models are trained identically to the Tacotron model. The model is trained according to the L1 loss between the target spectrogram and the generated spectrogram, and no other supervision is given for the reference encoder. Unless otherwise stated, we used the same hyperparameter settings used in earlier work \cite{Prosody}.

\subsection{Speech-side prosody control}
\label{ssec:speech_side}

The length $l_p$ of variable-length prosody embedding created from a reference spectrogram with length $l_{ref}$ is identical to $l_{ref}$. Note that the decoder should generate the same spectrogram as the reference spectrogram and that $r$-frames are generated at each decoder time-step. This gives $l_p$ a longer length by $r$-times than $l_d$. By choosing appropriate stride sizes for the convolutional layers, we could shorten reference spectrogram to match $l_p$ with $l_d$.

At each decoder time-step $i$, $p_i$ is initially fed to the attention module together with $e_{1:l_{e}}$ to compute the $i$-th attention weights, $\alpha_i$. 
We did not feed speaker embedding to the attention module as we assumed the speaker identity to be conditionally independent with attention weights when prosody is given. 
The weighted sum of $e_{1:l_{e}}$ with $\alpha_i$ gives us the context vector $e'_i$. The input of the decoder module at the $i$-th time-step is a concatenation of $\{e'_i, p_i, s\}$. 
\begin{equation}\label{eq:s_side}
\begin{aligned}
    e_{1:l_e} &= \text{Encoder}(x_{1:l_e})\\
    \alpha_i &= \text{Attention}(e_{1:l_{e}}, p_i, d_{i-1})\\
    e'_i &= \Sigma_j \alpha_{ij} e_j\\
    d_i &= \text{Decoder}(e'_i, p_i, s)
\end{aligned}
\end{equation}

\begin{figure*}[t]
\centering
\begin{minipage}[b]{.2\linewidth}
  \centering
  \centerline{\includegraphics[width=\textwidth]{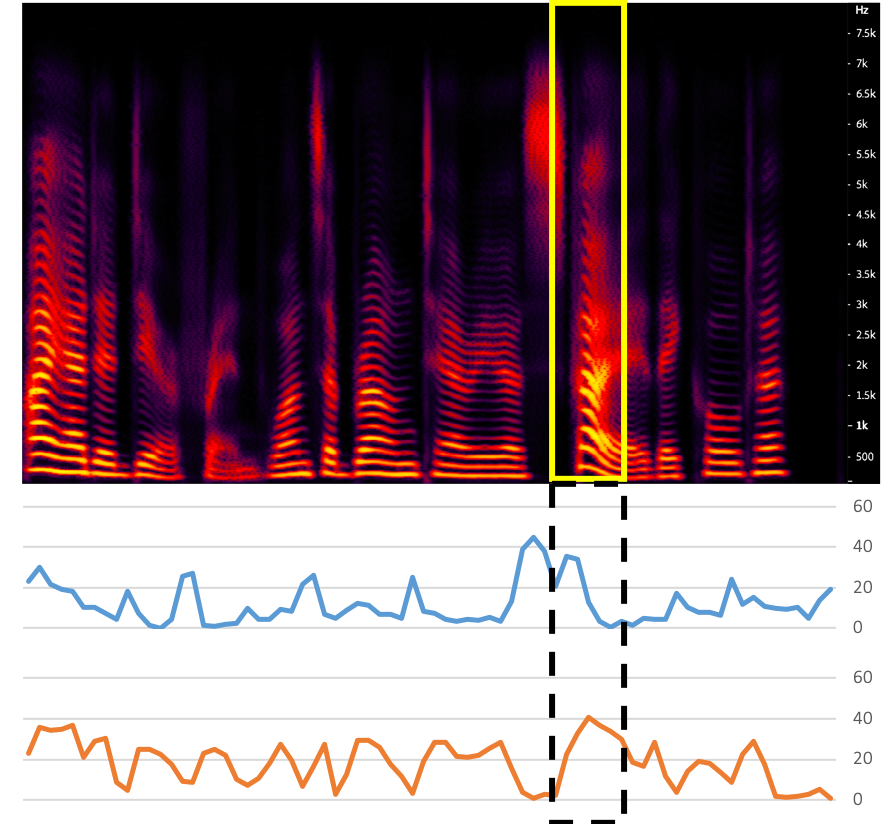}}
  \centerline{(a) Original prosody}\medskip
\end{minipage}
\hfill
\begin{minipage}[b]{0.2\linewidth}
  \centering
  \centerline{\includegraphics[width=\textwidth]{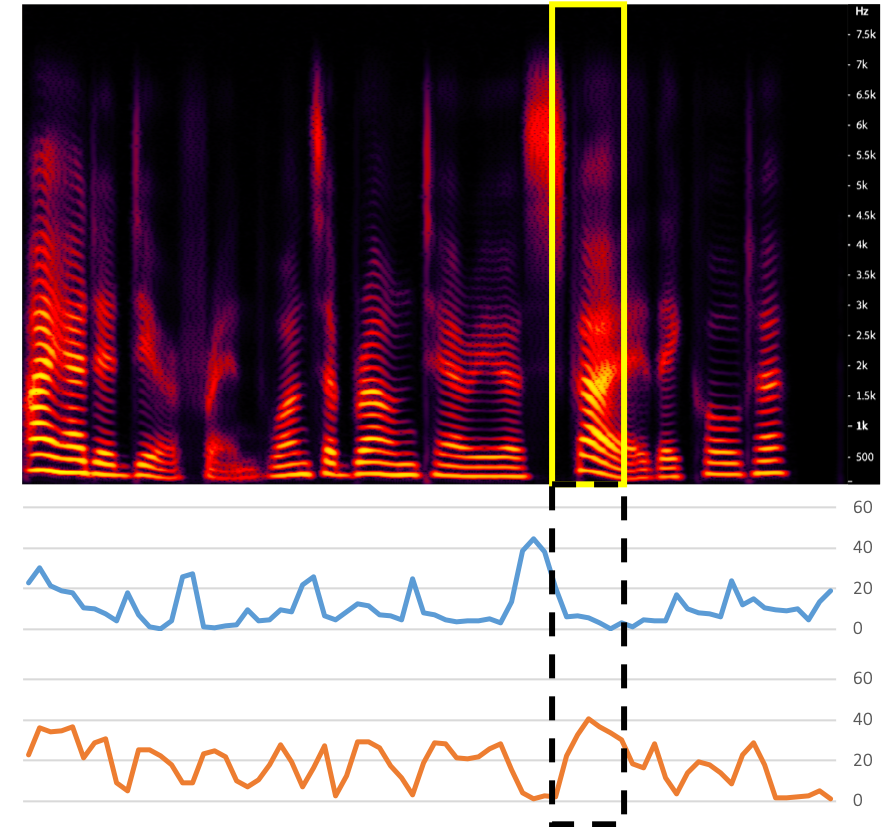}}
  \centerline{(b) Adjusted 1st prosody}\medskip
\end{minipage}
\hfill
\begin{minipage}[b]{0.2\linewidth}
  \centering
  \centerline{\includegraphics[width=\textwidth]{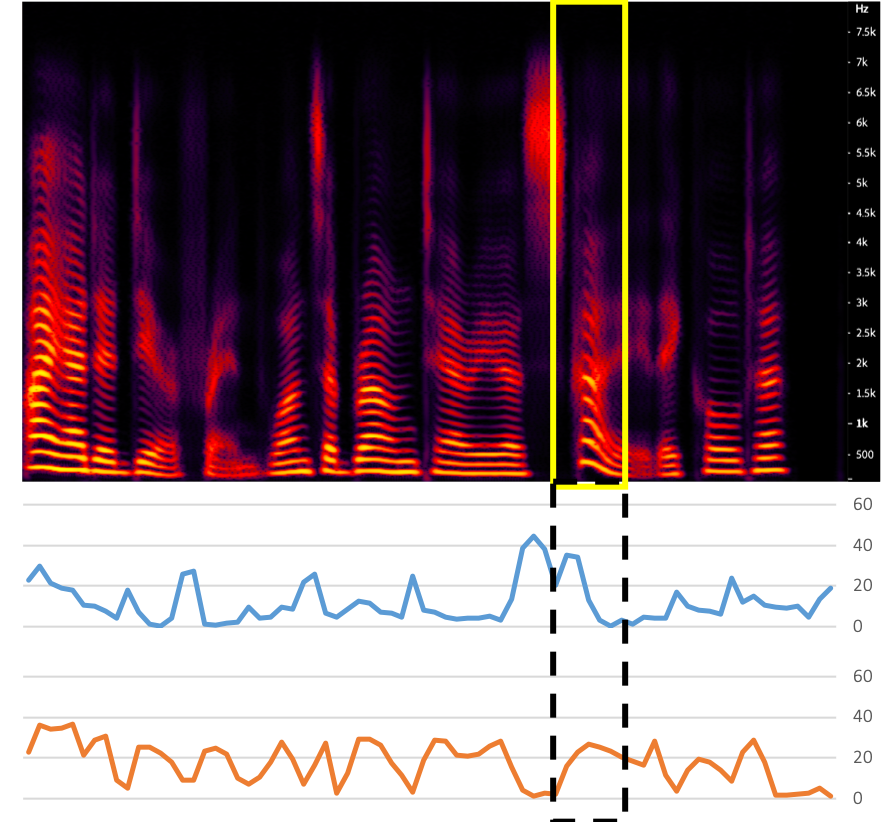}}
  \centerline{(c) Adjusted 2nd prosody}\medskip
\end{minipage}
\hfill
\begin{minipage}[b]{0.2\linewidth}
  \centering
  \centerline{\includegraphics[width=\textwidth]{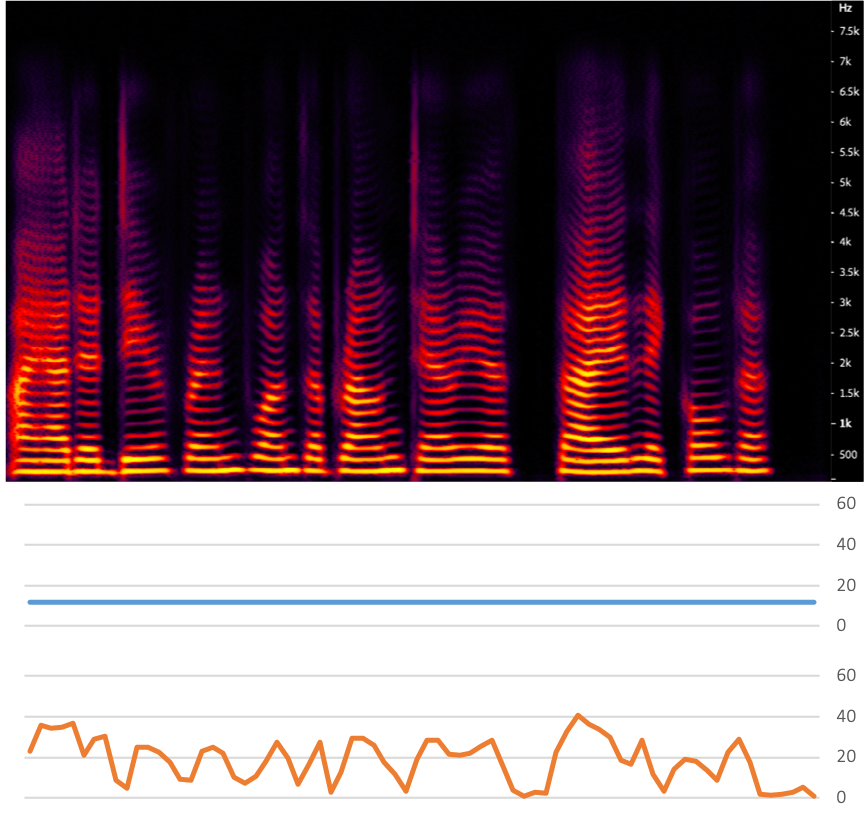}}
  \centerline{(d) Fixed 1st prosody}\medskip
\end{minipage}
\caption{Speech-side prosody control.}
\label{fig:prosody-speech}
\end{figure*}

\begin{figure*}[t]
\centering
\begin{minipage}[b]{.19\linewidth}
  \centering
  \centerline{\includegraphics[width=\textwidth]{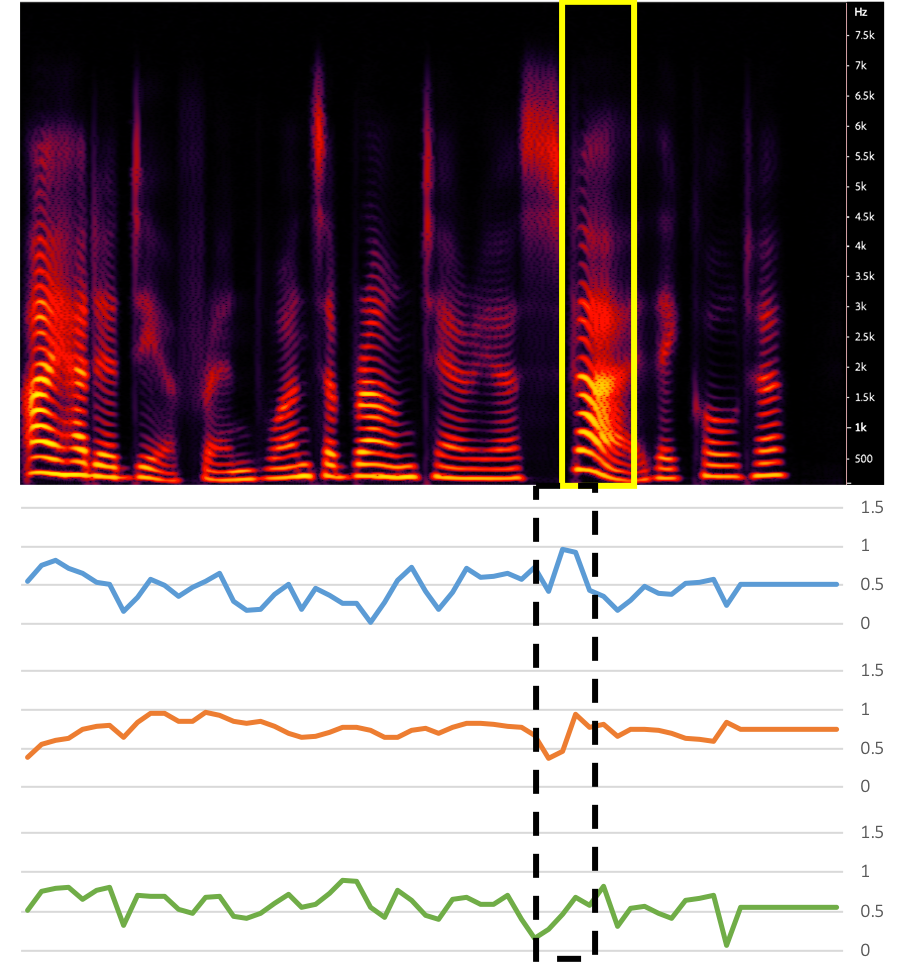}}
  \centerline{(a) Original prosody}\medskip
\end{minipage}
\hfill
\begin{minipage}[b]{0.19\linewidth}
  \centering
  \centerline{\includegraphics[width=\textwidth]{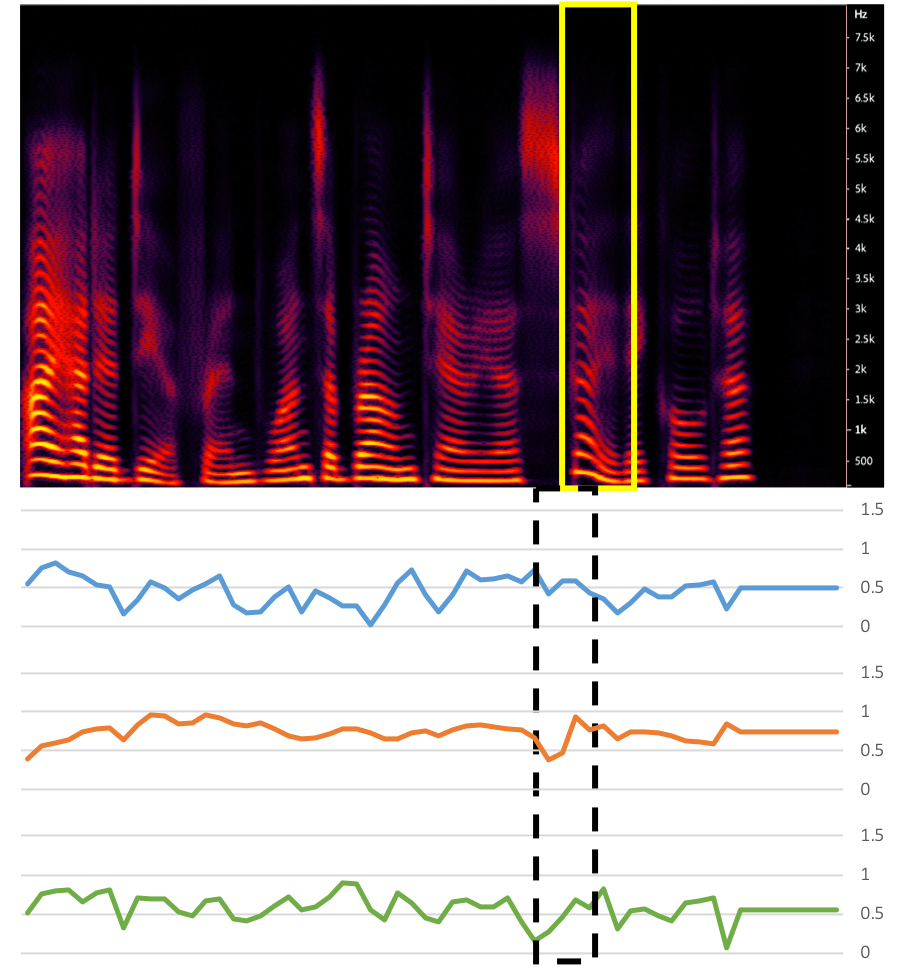}}
  \centerline{(b) Adjusted 1st prosody}\medskip
\end{minipage}
\hfill
\begin{minipage}[b]{0.19\linewidth}
  \centering
  \centerline{\includegraphics[width=\textwidth]{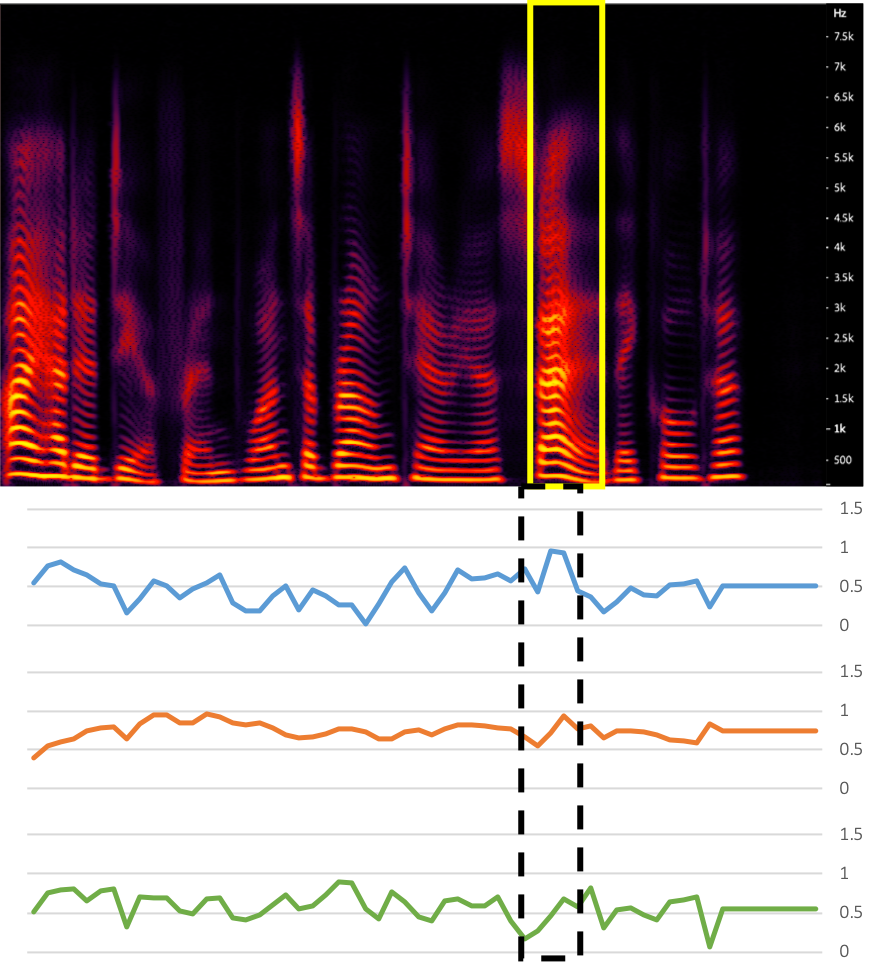}}
  \centerline{(c) Adjusted 2nd prosody}\medskip
\end{minipage}
\hfill
\begin{minipage}[b]{0.19\linewidth}
  \centering
  \centerline{\includegraphics[width=\textwidth]{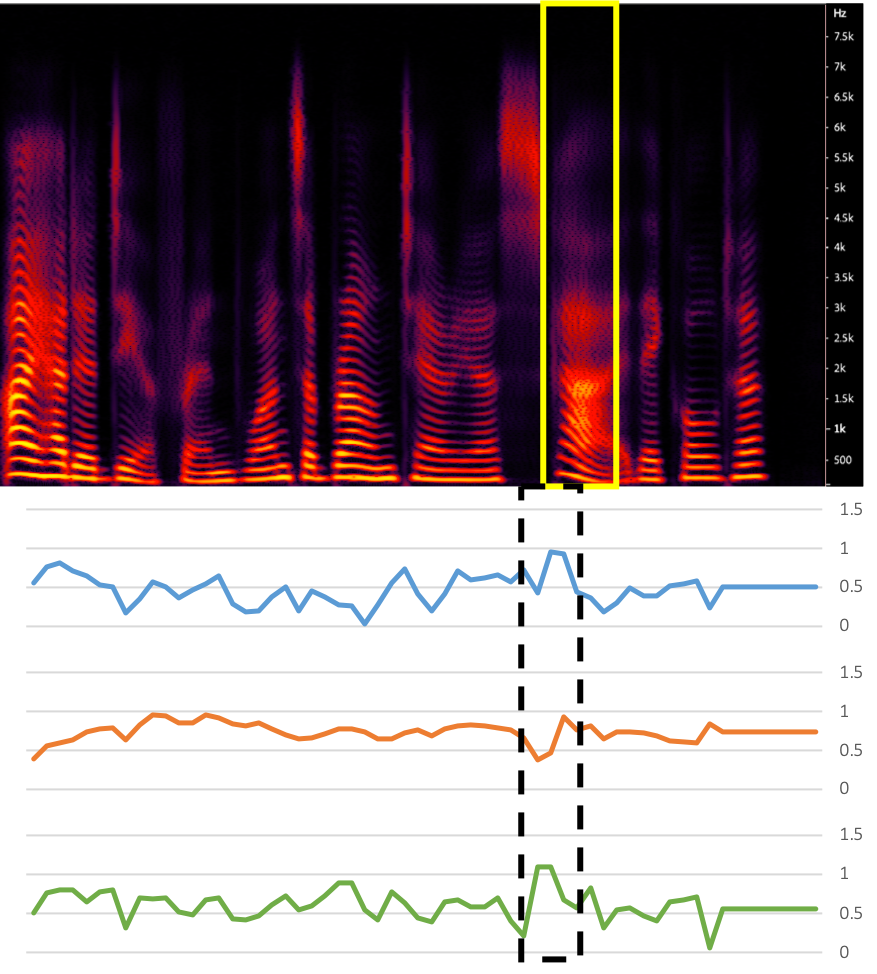}}
  \centerline{(d) Adjusted 3rd prosody}\medskip
\end{minipage}
\hfill
\begin{minipage}[b]{0.19\linewidth}
  \centering
  \centerline{\includegraphics[width=\textwidth]{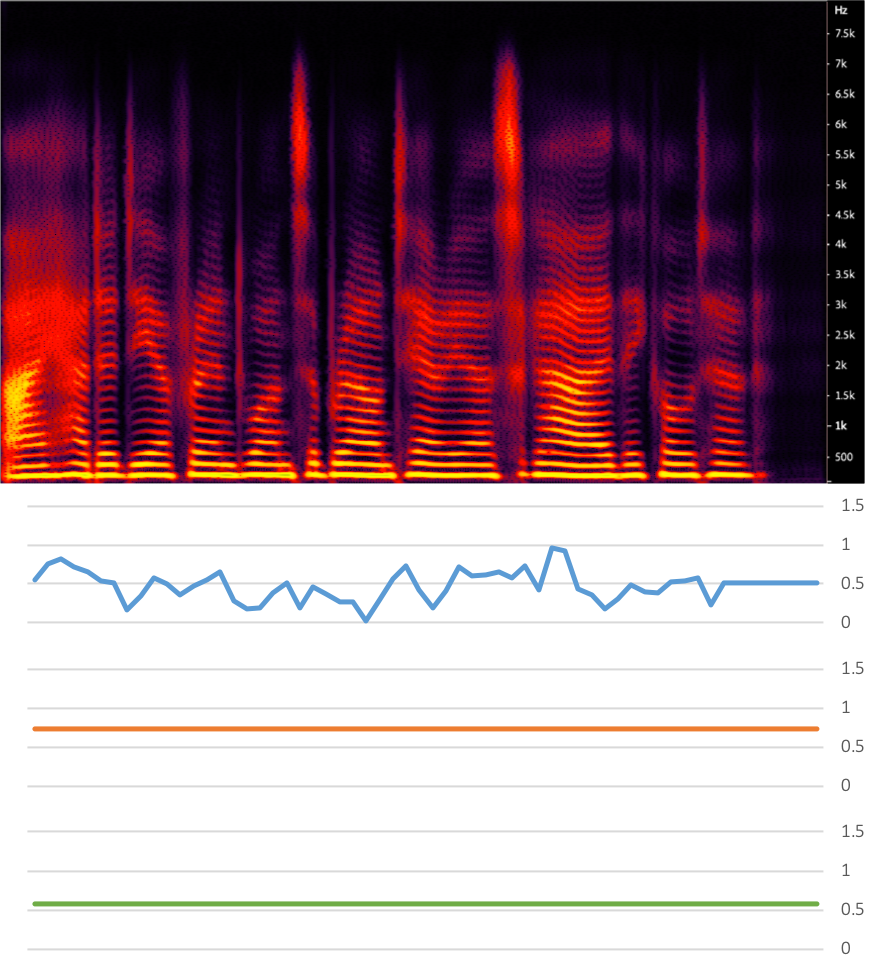}}
  \centerline{(e) Fixed 2nd/3rd prosody}\medskip
\end{minipage}
\caption{Text-side prosody control.}
\label{fig:prosody-text}
\end{figure*}

\subsection{Text-side prosody control}
\label{ssec:text_side}
The linear relationship between $l_p$ and $l_d$ made it easy to ensure that speech-side prosody embedding has a length identical to the number of the decoder time-steps. Unfortunately, such a relationship is not guaranteed between $l_p$ and $l_e$. We introduced a reference attention module that uses scaled dot-product attention \cite{Transformer} to find the alignment between $e_{1:l_{e}}$ and $p_{1:l_{ref}}$. In the reference attention module, \textit{key} $\kappa$ and \textit{value} $v$ are obtained from $p$ and the \textit{query} is $e$. Conceptually, the attention mechanism computes the attention weight according to the similarity between the \textit{query} and each \textit{key}, and weighted sum of the \textit{values} is then obtained using the attention weight. To obtain $\kappa$ and $v$ from prosody embedding, we doubled the output dimension $h$ of the reference encoder for the text-side prosody control, with the output split into two matrices of size ($l_{ref} \times h$). The weighted sum of $v_{1:l_{ref}}$ with $\beta$ gives us text-side prosody embedding $p^t$. Then, $p^t$ is concatenated to $e$ upon every usage of $e$.

\begin{equation}\label{eq:t_side}
\begin{aligned}
    e_{1:l_e} &= \text{Encoder}(x_{1:l_e})\\
    \big[\kappa_{1:l_{ref}}; v_{1:l_{ref}} \big] &= p_{1:l_{ref}}\\
    \beta_j &= \text{Ref-Attention}(e_j, \kappa_{1:l_{ref}})\\
    p^t_j &= \Sigma_{k} \beta_{jk} v_k\\
    \alpha_i &= \text{Attention}(\big[e_{1:l_{e}}; p^t_{1:l_{e}}\big], d_{i-1})\\
    e'_i &= \Sigma_{j} \alpha_{ij} \big[e_j; p^t_j\big]\\
    d_i &= \text{Decoder}(e'_i, s)
\end{aligned}
\end{equation}

\subsection{Prosody normalization}
\label{ssec:normalization}

Prosody embedding is normalized using each speaker's prosody mean. During training, we computed the sample mean along the temporal dimension of variable-length prosody embedding and stored average of the sample mean for each speaker. For both the training step and the evaluation, normalization was done by subtracting the speaker-wise prosody mean from every time step of prosody embedding.

\section{Experiments and results}
\label{sec:experiment}

\subsection{Dataset}
\label{ssec:data}

Previous works \cite{Prosody, SToken} used large amounts of data to train the prosodic TTS model (296 hours of data for the multi-speaker model). To ensure a large amount of data, we used multiple datasets, in this case VCTK, CMU ARCTIC, and internal datasets. The final dataset consisted of 104 hours (58 hours of English and 46 hours of Korean) with 136 speakers (128 English speakers and 8 Korean speakers).

Because variable-length prosody embedding has a large enough capacity to copy the reference audio, we had to use a very small dimension for the bottleneck size. This led us to the use of prosody sizes of 2 and 4 for the speech-side and text-side prosody embedding, respectively.

\subsection{Speech-side control of prosody}
\label{ssec:exp_s_side}
By adjusting the values of the speech-side prosody embedding, we could change the prosody at a specific frame. Figure \ref{fig:prosody-speech} shows the change in the learned prosody embeddings (line graph) and their corresponding spectrograms. The first dimension of prosody embedding, in the second row of Figure \ref{fig:prosody-speech}, tended to control the pitch of the generated speech. By comparing the highlighted parts of Figures \ref{fig:prosody-speech}-(a) and (b), one can assess the change of the pitch from the spaces between the harmonics. The second dimension of prosody embedding, in the third row of Figure \ref{fig:prosody-speech}, tended to control the amplitude of the generated speech. By comparing the highlighted parts of Figures \ref{fig:prosody-speech}-(a) and (c), one can assess the change of the amplitude from the intensity of the harmonics. We recommend that readers listen to the examples on our demo page.\footnote{http://neosapience.com/research/2018/10/29/icassp}


\begin{figure}[t]
\begin{minipage}[b]{.48\linewidth}
  \centering
  \centerline{\includegraphics[width=4.0cm]{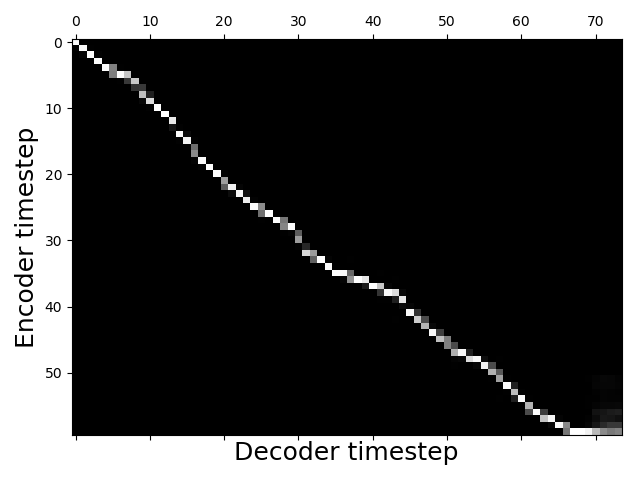}}
  \centerline{(a) Original attention}\medskip
\end{minipage}
\hfill
\begin{minipage}[b]{0.48\linewidth}
  \centering
  \centerline{\includegraphics[width=4.0cm]{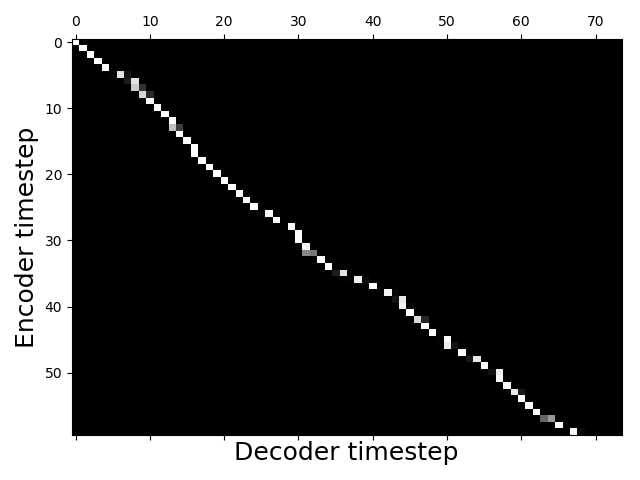}}
  \centerline{(b) Reference attention}\medskip
\end{minipage}
\caption{Attention alignment between text and speech}
\label{fig:att_plot}
\end{figure}

\begin{figure*}[t]
\centering
\begin{minipage}[b]{.22\linewidth}
  \centering
  \centerline{\includegraphics[width=\textwidth]{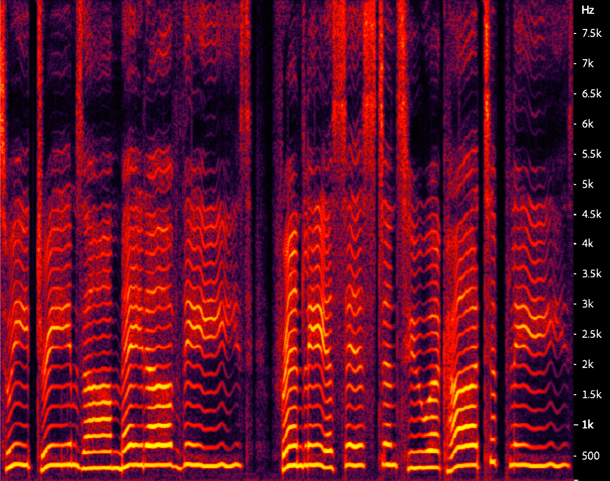}}
  \centerline{(a) Original song}\medskip
\end{minipage}
\hfill
\begin{minipage}[b]{0.22\linewidth}
  \centering
  \centerline{\includegraphics[width=\textwidth]{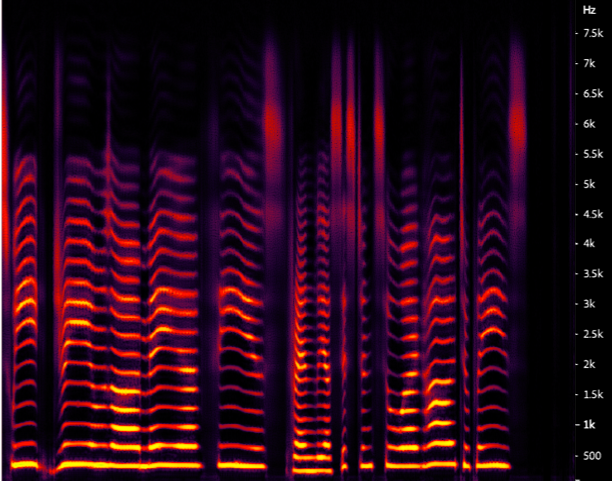}}
  \centerline{(b) GST}\medskip
\end{minipage}
\hfill
\begin{minipage}[b]{0.22\linewidth}
  \centering
  \centerline{\includegraphics[width=\textwidth]{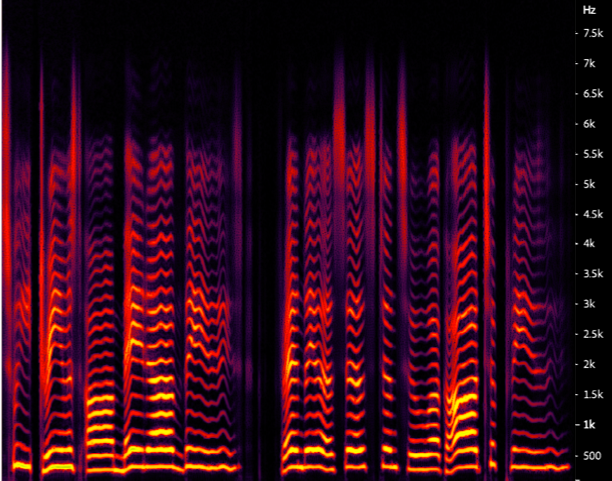}}
  \centerline{(c) Speech-side prosody}\medskip
\end{minipage}
\hfill
\begin{minipage}[b]{0.22\linewidth}
  \centering
  \centerline{\includegraphics[width=\textwidth]{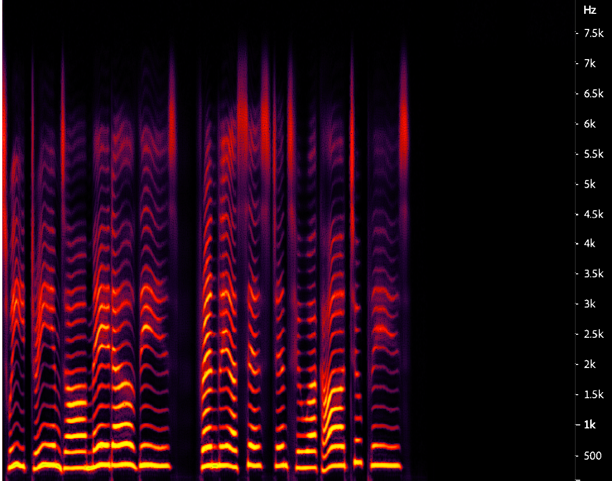}}
  \centerline{(d) Text-side prosody}\medskip
\end{minipage}
\caption{Spectrogram from singing voice transfer.}
\label{fig:prosody-sing}
\end{figure*}

\subsection{Text-side control of prosody}
\label{ssec:exp_t_side}
First, we checked if the reference attention module learned how to find the alignment between the phoneme sequence and the reference audio. Figure \ref{fig:att_plot} shows an attention alignment plot of the original attention module (a) and reference attention module (b). From their analogous shape, we find that the reference attention module could align the reference speech to the text. 

As was done in Section \ref{ssec:exp_s_side}, we changed the prosody of the phonemes by adjusting the text-side prosody embedding in Figure \ref{fig:prosody-text}. It appeared that the amplitude was affected by the first and third dimensions and that the pitch was affected by the second and third dimensions. In addition, the length was affected by the first and third dimensions. It would be ideal if each dimension represents one prosodic feature (i.e., the pitch, amplitude, or length). We think prosody embedding is entangled because we did not impose any constraints on prosody embedding to be disentangled.

\subsection{Comparison with GST Tacotron}
\label{ssec:comparison}
We compared our methods to GST Tacotron both quantitatively and qualitatively. For the quantitative comparison, we used the Mean Cepstral Distortion (MCD) with the first 13 MFCCs, as proposed in earlier work \cite{Prosody}. Table \ref{table-mcd} shows that the proposed methods outperform GST Tacotron in terms of MCD$_{13}$, where a lower MCD is better. In particular, speech-side prosody control, which has the highest temporal resolution of prosody embedding, showed the lowest MCD.

\begin{table}[ht]
\centering
\caption{Mean cepstral distortion of types of prosody embedding}
\label{table-mcd}
\resizebox{0.3\textwidth}{!}{%
\begin{tabular}{|c|c|}
\hline
Model & MCD$_{13}$ \\ \hline
Global style token & 0.413 \\ \hline
Speech-side prosody control & 0.294 \\ \hline
Text-side prosody control & 0.342 \\ \hline
\end{tabular}%
}
\end{table}

One shortcoming of GST Tacotron is that GST works only in a global sense. If we fix GST for multiple decoder time steps, the decoder generates speech while changing the prosody implicitly at each time step to create the GST's speech style. This is not problematic if the generated prosody perfectly matches the intention of the user, but in many cases we needed modifications to realize this. Because GST changes the prosody implicitly, it is ambiguous to control the prosody at specific moment. On the other hand, the proposed prosody embeddings control the prosody explicitly. In Sections \ref{ssec:exp_s_side} and \ref{ssec:exp_t_side}, we observed that prosody can be controlled by adjusting the values of the prosody embedding. We further demonstrated the explicitness and consistency of the proposed methods by fixing prosody embedding to have the same value over all frames. This should give us a flat speech style in contrast to the GST approach, and we can see these outcomes in Figure \ref{fig:prosody-speech}-(d) and Figure \ref{fig:prosody-text}-(e). The results are obtained by fixing the dimensions that controlled the pitch.

If we apply prosody control to a wider dynamic range of prosody embedding, it will be able to generate a singing voice. We demonstrate this in Figure \ref{fig:prosody-sing}. Using the three prosody control methods, we extracted the prosody from an unseen song of an unseen singer. We combined the extracted prosody embedding with the original lyrics and a speaker in the training set to perform the prosody transfer. While GST could not reconstruct the melody of the song, we could recognize the melody of the original song using the proposed methods. In particular, the generated song from speech-side prosody control was almost identical to the original song. For this task, the lyrics, speaker identity, and prosody embedding were the only requirements for the generation step. We witnessed the capability of the proposed methods to generate a song given an appropriate sequence of prosody embedding.

\subsection{Inter-speaker prosody transfer}
\label{ssec:exp_transfer}
We compared the MCD of the speech-side prosody embeddings with and without normalization, as shown in Table \ref{table-mcd-normal}. For each case of prosody embedding, we computed the MCD between the reference and the generated speech for each prosody reconstruction and prosody transfer task. In both tasks, we used the speech of a female speaker as the reference speech, and we used the speaker ID of the same female speaker or another male speaker for prosody reconstruction and prosody transfer, respectively. Without normalization, the generated speech tended to show a higher pitch than the male speaker and sometimes failed to generate speech. We consider that this failure arises because the combination of the male speaker ID and female prosody embedding did not exist during the training step. When we used normalization for prosody embedding, the model was exposed to the similarly distributed prosody embedding during the traininng phase. This caused the prosody transfer to be easier compared to that without normalization. Table \ref{table-mcd-normal} also presents this phenomenon with a higher MCD during the prosody transfer with the non-normalized model compared to that with the normalized model.

\begin{table}[ht]
\centering
\caption{Mean cepstral distortion of types of prosody transfer}
\label{table-mcd-normal}
\resizebox{0.4\textwidth}{!}{%
\begin{tabular}{|c|c|c|}
\hline
Model & Female-to-Female & Female-to-Male \\ \hline
Normalized & 0.329 &  0.518 \\ \hline
Not-normalized & 0.304 & 0.531 \\ \hline
\end{tabular}%
}
\end{table}

\section{CONCLUSION AND FUTURE WORK}
\label{sec:conclusion}
Here, we proposed temporally structured prosody embedding networks to control the expressive style of synthesized speech. The proposed methods changed the pitch and amplitude both at the frame-level and phoneme-level resolution. Moreover, normalized prosody embedding made the prosody transfer step more robust to pitch discrepancies between the reference and generated speaker. The proposed methods demonstrated better quality in terms of the MCD score, and the prosody of a song could be successfully transferred to another speaker, resulting in voice conversion of a song.

The bottleneck size was the only factor that regularized the prosody embedding network in this paper. Disentangling techniques will be beneficial to factorize the prosody embeddings into more explainable prosodic features and separate them from other speech features. This will be a fruitful direction for future work.


\vfill\pagebreak

\bibliographystyle{IEEEbib}
\bibliography{refs}

\end{document}